\documentclass[twoside]{article}
\usepackage[accepted]{icml2013} 

\usepackage{graphicx} 
\usepackage{natbib}
\usepackage{algorithm}
\usepackage{algorithmic}
\usepackage{amssymb}
\usepackage{amsmath}
\def\href#1#2{#1}
\usepackage{array}
\usepackage{multirow}

\newenvironment{proof}[1][Proof]{\begin{trivlist}
\item[\hskip \labelsep {\bfseries #1}]}{\end{trivlist}}

\newcommand{\qed}{\hfill \ensuremath{\Box}}
\def\slantfrac#1#2{\kern.1em^{#1}\kern-.3em/\kern-.1em_{#2}}

\def\nabla{\bigtriangledown}
\newcommand{\inp}{\mathbf{u}}

\newcommand{\loss}{\mathcal{E}}
\newcommand{\state}{\mathbf{x}}
\makeatletter
\newcommand\footnoteref[1]{\protected@xdef\@thefnmark{\ref{#1}}\@footnotemark}
\makeatother

\icmltitlerunning{On the difficulty of training Recurrent Neural Networks}
\begin{document}
\twocolumn[

\icmltitle{On the difficulty of training Recurrent Neural Networks}

\icmlauthor{Razvan Pascanu}{pascanur@iro.umontreal.ca}
\icmladdress{Universite de Montreal}
\icmlauthor{Tomas Mikolov}{t.mikolov@gmail.com}
\icmladdress{Brno University}
\icmlauthor{Yoshua Bengio}{yoshua.bengio@umontreal.ca}
\icmladdress{Universite de Montreal}
\icmlkeywords{recurrent neural networks, vanishing gradients, exploding gradients}
]

\begin{abstract}
There are two widely known issues with properly training Recurrent Neural 
Networks, the \emph{vanishing} and the \emph{exploding} gradient problems 
detailed in \citet{Bengio-trnn94}. In this paper we attempt
to improve the understanding of the underlying issues by exploring these 
problems
from an analytical, a geometric and a dynamical systems perspective. 
Our analysis is used to justify a simple yet effective solution. 
We propose a gradient norm clipping strategy to deal with
exploding gradients and a soft constraint for the vanishing 
gradients problem. 
We validate empirically our hypothesis and proposed solutions 
in the experimental section.
\end{abstract}

\vspace*{-3mm}
\section{Introduction}
\label{sec:intro}
\vspace*{-1mm}

\begin{figure}[ht]
\begin{center}
\centerline{\includegraphics[width=.3\columnwidth]{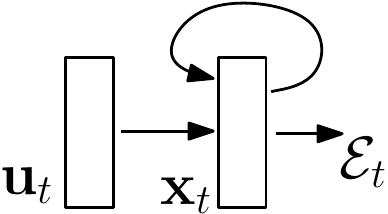}}
\caption{Schematic of a recurrent neural network. The recurrent 
    connections in the hidden layer allow information to persist 
    from one input to another.} 
\label{fig:RNN}
\end{center}
\end{figure}

A recurrent neural network (RNN), e.g. Fig. \ref{fig:RNN}, is a 
neural network model proposed in the 80's~\citep{BP86, Elman90, 
Werbos88} for modeling time series. The structure 
of the network is similar to that of a standard multilayer perceptron,
with the distinction that we allow connections among hidden units 
associated with a time delay. Through these connections the model 
can retain information about the past inputs, enabling it 
to discover temporal correlations between events that are possibly 
far away from each other in the data (a crucial property for 
proper learning of time series). 

While in principle the recurrent network is a simple and powerful model, 
 in practice, it is unfortunately hard to train properly.
Among the main reasons why this 
model is so unwieldy are the \emph{vanishing gradient} and \emph{exploding 
gradient} problems described in \citet{Bengio-trnn94}.


\vspace*{-2mm}
\subsection{Training recurrent networks}
\label{sec:problem}
\vspace*{-1mm}

A generic recurrent neural network, with input $\inp_t$ and state $\state_t$
for time step $t$, is given by equation \eqref{eq:generic_rnn}.
In the theoretical section of this paper we will sometimes make use 
of the specific parametrization given by equation \eqref{eq:standard_rnn}
\footnote{
    This formulation is equivalent to the more widely known equation 
    $\state_{t} = \sigma(\mathbf{W}_{rec} \state_{t-1} + \mathbf{W}_{in} \inp_t + \mathbf{b})$, 
    and it was chosen for convenience.} 
in order to provide more precise conditions and intuitions about the everyday use-case. 
\begin{equation}
\label{eq:generic_rnn}
\state_t = F(\state_{t-1}, \inp_{t}, \theta)
\end{equation}
\begin{equation}
    \label{eq:standard_rnn}
    \state_{t} = \mathbf{W}_{rec} \sigma(\state_{t-1}) + \mathbf{W}_{in} \inp_{t} + \mathbf{b}
\end{equation}
The parameters of the model are given by the recurrent weight matrix $\mathbf{W}_{rec}$,
the biases $\mathbf{b}$ 
and input weight matrix $\mathbf{W}_{in}$, collected in $\theta$ for the general case. 
$\state_{0}$ is provided by the user, set to zero or learned,
and $\sigma$ is an element-wise function (usually the \textit{tanh} or \textit{sigmoid}). 
A cost $\loss$ measures the performance of the network on some 
given task and it can be broken apart into individual costs for 
each step $\loss = \sum_{1 \leq t \leq T}\loss_t$, where 
$\loss_t = \mathcal{L}(\state_{t})$. 

\begin{figure}[ht]
\vskip 0.2in
\begin{center}
    \centerline{
    \includegraphics[width=\columnwidth]{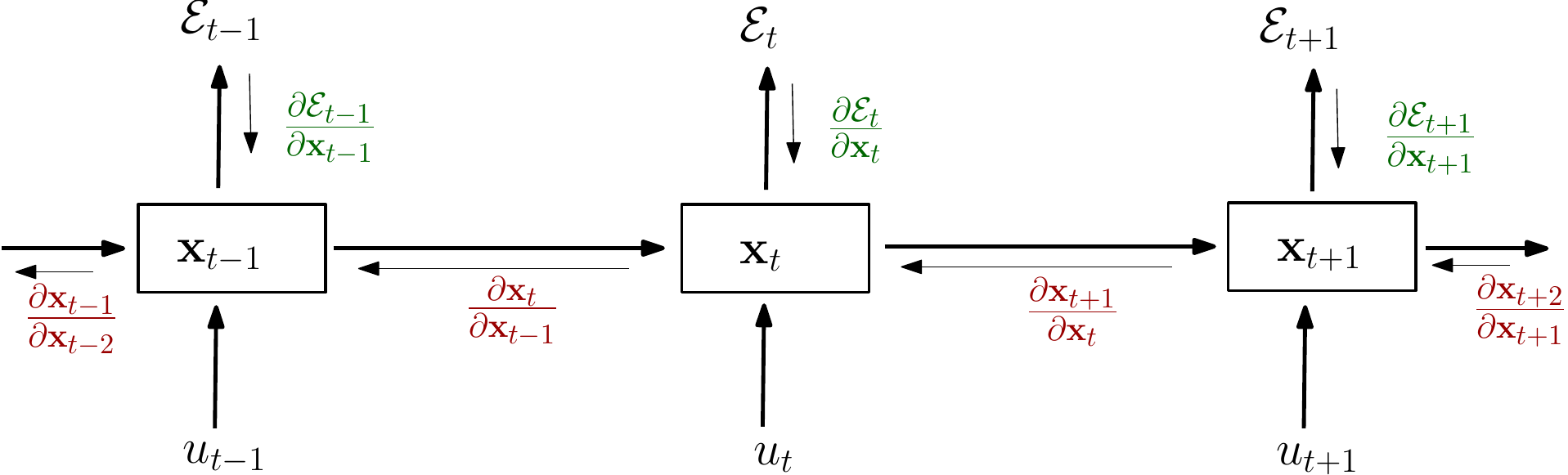}}
  \caption{Unrolling recurrent neural networks in time by creating 
      a copy of the model for each time step. We denote by 
      $\mathbf{x}_t$ the hidden state of the network at time $t$, 
      by $\mathbf{u}_t$ the input of the network at time $t$ 
      and by $\loss_t$ the error obtained from the output at 
      time $t$. } 
  \label{fig:unrolled}
  \end{center}
  \vskip -0.2in
\end{figure}

One approach that can be used to compute the necessary gradients is Backpropagation 
Through Time (BPTT), where the recurrent model 
is represented as a deep 
multi-layer one (with an unbounded number of layers) and backpropagation is applied
on the unrolled model (see Fig. \ref{fig:unrolled}). 

We will diverge from the classical BPTT equations at this point and 
re-write the gradients (see equations \eqref{eq:grad_theta1}, 
\eqref{eq:grad_theta2} and \eqref{eq:grad_theta3}) in order to better 
highlight the exploding gradients problem. These equations were obtained 
by writing the gradients in a sum-of-products form.
\begin{equation}
\label{eq:grad_theta1}
\frac{\partial \loss}{\partial \theta} = 
\sum_{1 \leq t \leq T} \frac{\partial \loss_t}{\partial \theta}
\end{equation}
\begin{equation}
\label{eq:grad_theta2}
\frac{\partial \loss_t}{\partial \theta} = 
\sum_{1 \leq k \leq t} 
\left(
\frac{\partial \loss_t}{\partial \state_{t}}
\frac{\partial \state_{t}}{\partial \state_{k}}
\frac{\partial^+ \state{_{k}}}{\partial \theta}
\right)
\end{equation}
\begin{equation}
\label{eq:grad_theta3}
\frac{\partial \state_{t}}{\partial \state_{k}}=
\prod_{t \geq i > k} \frac{\partial \state_i}{\partial \state_{i-1}} =
\prod_{t \geq i > k} \mathbf{W}_{rec}^T diag (\sigma'(\state_{i-1}))
\end{equation}
$\frac{\partial^+ \state{_{k}}}{\partial \theta}$ 
refers to the ``immediate''
partial derivative of the state $\mathbf{x}_k$ with respect 
to $\theta$, i.e., where $\mathbf{x}_{k-1}$ is taken as a constant with respect to $\theta$. 
Specifically, considering equation 2, the value of any row $i$ of the matrix
$(\frac{\partial^+ \mathbf{x}_k}{\partial \mathbf{W}_{rec}})$ is just $\sigma(\mathbf{x}_{k-1})$.
Equation \eqref{eq:grad_theta3} 
also provides the  form of Jacobian matrix $\frac{\partial \state_i}{\partial  \state_{i-1}}$ 
for the specific parametrization given in equation 
\eqref{eq:standard_rnn}, where $diag$ converts a vector into a diagonal matrix, 
and $\sigma'$ computes the derivative of $\sigma$ in an element-wise
fashion. 

Note that each term $\frac{\partial \loss_t}{\partial \theta}$ 
from equation \eqref{eq:grad_theta1} has the same form and the 
behaviour of these individual terms determine the behaviour of the sum. 
Henceforth we will focus on one such generic term, calling 
it simply the gradient when there is no confusion.

Any gradient component $\frac{\partial \loss_t}{\partial \theta}$ 
is also a sum (see equation \eqref{eq:grad_theta2}), whose terms we 
refer to as \emph{temporal} contributions or \emph{temporal} 
components. One can see that each such temporal contribution 
$\frac{\partial \loss_t}{\partial \state_{t}}
\frac{\partial \state_{t}}{\partial \state_{k}}
\frac{\partial^+ \state{_{k}}}{\partial \theta}
$ measures how $\theta$ at step $k$ affects the cost at step $t>k$. 
The factors $\frac{\partial \state_{t}}{\partial \state_{k}}$ 
(equation \eqref{eq:grad_theta3}) transport the error ``in time`` 
from step $t$ back to step $k$. We would further loosely distinguish 
between \emph{long term}  
and \emph{short term} contributions, 
where long term refers 
to components for which $k \ll t$ and short term to everything 
else.

\vspace*{-2mm}
\section{Exploding and Vanishing Gradients}
\label{sec:theoretical}
\vspace*{-.5mm}

As introduced in \citet{Bengio-trnn94}, the \emph{exploding gradients} problem
refers to the large increase in the norm of the gradient during training.
Such events are caused by the explosion of the long term components, 
which can grow exponentially more then short term ones.
The \emph{vanishing gradients} problem refers to the opposite behaviour, when 
long term components go exponentially fast to norm 0, making it impossible 
for the model to learn correlation between temporally distant events.

\vspace*{-2mm}
\subsection{The mechanics}
\label{sec:mechanics}
\vspace*{-.5mm}

To understand this phenomenon we need to look at the form 
of each temporal component, and in particular at
the matrix factors
$\frac{\partial \state_t}{\partial \state_k}$ 
(see equation \eqref{eq:grad_theta3}) that take the form of a product of $t-k$ Jacobian matrices.
{\em In the same way a product of $t-k$ real numbers can shrink to zero or explode 
to infinity, so does this product of matrices} (along some direction $\mathbf{v}$). 

In what follows we will try to formalize these intuitions (extending 
a similar derivation done in \citet{Bengio-trnn94} where only a 
single hidden unit case was considered). 

If we consider a linear version of the model (i.e. set $\sigma$ to the identity function in 
equation \eqref{eq:standard_rnn}) we can use the \emph{power iteration method} to formally analyze 
this product of Jacobian matrices and obtain tight conditions for when the gradients 
explode or vanish (see the supplementary materials for a detailed derivation of these 
conditions). It is \emph{sufficient} for the largest eigenvalue $\lambda_1$
of the recurrent weight matrix to be 
smaller than 1 for long term components to vanish (as $t \to \infty$) and \emph{necessary} for it to be 
larger than 1 for gradients to explode.

We can generalize these results for nonlinear functions $\sigma$ where the absolute values of $\sigma'(x)$ 
is bounded (say by a value $\gamma \in \mathcal{R}$) and therefore $\left\|diag(\sigma'(\state_k))\right\| \leq \gamma$.

We first {\bf prove} that it is \emph{sufficient} for $\lambda_1 < \frac{1}{\gamma}$, 
where $\lambda_1$ is the absolute value of the largest eigenvalue 
of the recurrent weight matrix $\mathbf{W}_{rec}$, for the \emph{vanishing gradient} problem to occur.
Note that we assume the parametrization given by equation \eqref{eq:standard_rnn}. The Jacobian matrix
$\frac{\partial \state_{k+1}}{\partial \state_k}$ is given by $\mathbf{W}_{rec}^T {diag}(\sigma'(\state_k))$. The 2-norm of this Jacobian is bounded by the 
product of the norms of the two matrices (see equation 
\eqref{eq:norm_bounding}). Due to our assumption, this implies that it is smaller than 1. 
\begin{equation}
\label{eq:norm_bounding}
\forall k, \left\|\frac{\partial \state_{k+1}}{\partial \state_{k}}\right\| \leq 
\left\|\mathbf{W}_{rec}^T\right\|\left\|{diag}(\sigma'(\state_k))\right\| < \frac{1}{\gamma}\gamma < 1
\end{equation}
Let $\eta \in \mathrm{R}$ be such that $\forall k, \left\|\frac{\partial \state_{k+1}}{\partial \state_{k}}\right\| \leq \eta <1$. 
The existence of $\eta$ is given by equation \eqref{eq:norm_bounding}. 
By induction over $i$, we can show that
\begin{equation}
\label{eq:exp_shrink}
\frac{\partial \loss_t}{\partial \state_t} \left(\prod_{i=k}^{t-1} \frac{\partial \state_{i+1}}{\partial \state_i}\right) \leq \eta^{t-k} \frac{\partial \loss_t}{\partial \state_t} 
\end{equation}
As $\eta < 1$, it follows that, according to equation \eqref{eq:exp_shrink}, long term contributions (for which $t-k$ is large) go to 0 exponentially fast with $t-k$. \qed 

By inverting this proof we get the \emph{necessary} condition for \emph{exploding gradients}, namely
that the largest eigenvalue $\lambda_1$ is larger than $\frac{1}{\gamma}$ (otherwise the long term 
components would vanish instead of exploding). For $\tanh$ we have $\gamma=1$ while for sigmoid 
we have $\gamma = \slantfrac{1}{4}$. 

\vspace*{-2mm}
\subsection{Drawing similarities with Dynamical Systems}
\vspace*{-1mm}
\label{sec:dyn}

We can improve our understanding of the exploding gradients and vanishing 
gradients problems by employing a dynamical systems perspective, as it was done before 
in \citet{Doya93, Bengio_icnn93}. 

We recommend reading \citet{strogatz} for a formal and detailed treatment of 
dynamical systems theory. 
For any parameter assignment $\theta$, depending on the initial state $\state_0$, 
the state $\state_t$ of an autonomous dynamical system converges, under the repeated application of the 
map $F$, to one of several possible different attractor states (e.g. 
point attractors, though other type of attractors exist). The model could also find itself 
in a chaotic regime, case in which some of the following observations may not hold, but
that is not treated in depth here.
Attractors describe the asymptotic behaviour of the model.
The state space is divided into basins of attraction,
one for each attractor. If the model is started in one basin of attraction, 
the model will converge to the corresponding attractor as $t$ grows. 

Dynamical systems theory tells us that as $\theta$ changes slowly,
the asymptotic 
behaviour changes smoothly almost everywhere except for certain crucial points where 
drastic changes occur (the new asymptotic behaviour ceases to be topologically equivalent 
to the old one). These points are called bifurcation boundaries and 
are caused by attractors that appear, disappear or change shape.

\begin{figure}[ht]
\begin{center}
\centerline{\includegraphics[width=1.\columnwidth]{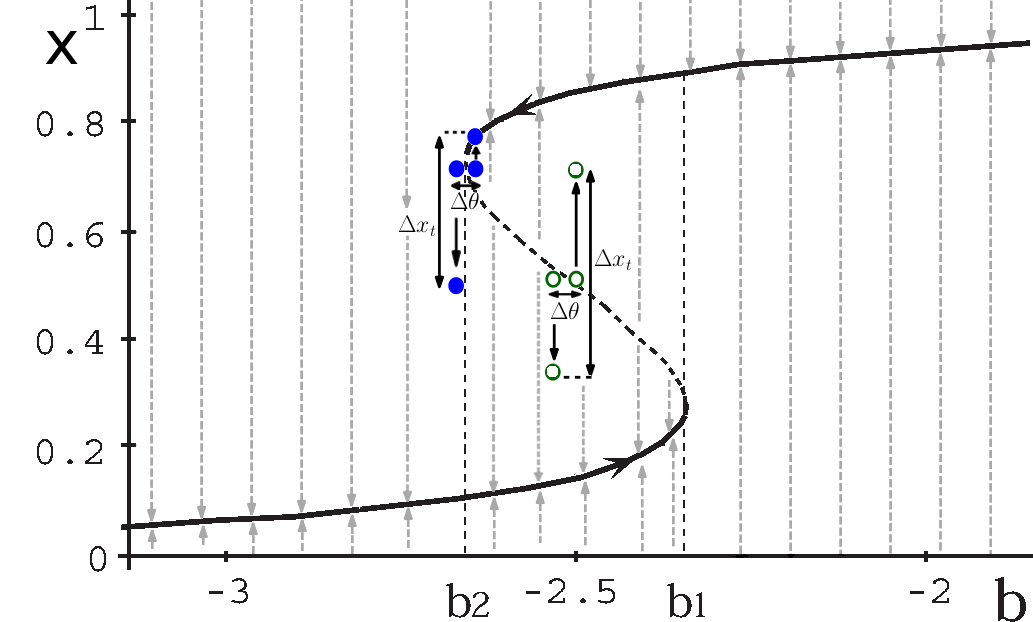}}
\vspace*{-3mm}
  \caption{Bifurcation diagram of a single hidden unit RNN (with fixed recurrent weight of 5.0 
  and adjustable bias $b$; example introduced in \citet{Doya93}). See text.} 
  \label{fig:state_space}
\end{center}
\vspace*{-5mm}
\end{figure}

\cite{Doya93} hypothesizes that such bifurcation crossings could cause the gradients 
to explode. We would like to extend this observation into a sufficient condition for gradients 
to explode, and for that reason we will re-use the one-hidden unit model (and plot) from 
\cite{Doya93} (see Fig.~\ref{fig:state_space}). 

The x-axis covers the parameter $b$ and the y-axis the asymptotic state $\state_\infty$.
The bold line follows the movement 
of the final point attractor, $\state_\infty$, as $b$ changes. At $b_1$ we have a bifurcation boundary where 
a new attractor emerges (when $b$ decreases from $\infty$), 
while at $b_2$ we have another that results in the disappearance of 
one of the two attractors. In the interval $(b_1, b_2)$ we are in a rich regime, 
where there are two attractors and the change in position of 
boundary between them, as we change $b$, 
is traced out by a dashed line. The vector field (gray dashed arrows) describe 
 the evolution of the state $\state$ if the network is initialized in that region.

We show that there are two types of events that could lead to a large change in $\state_t$, with $t\to\infty$. 
One is crossing a boundary between basins of attraction (depicted with a unfilled circles), while
the other is crossing a bifurcation boundary (filled circles).
For large $t$, the $\Delta x_t$ resulting from a change in $b$ will be large 
even for very small changes in $b$ (as the system is attracted towards different attractors) 
which leads to a large gradient. 

It is however \emph{neither necessary nor sufficient} 
to cross a  bifurcation for the gradients to explode,  
as bifurcations are global events that could have no effect locally. Learning traces out a 
path in the parameter-state space. If we are at a bifurcation boundary, but the state of the model 
is such that it is in the basin of attraction of one attractor (from many possible attractors) that 
does not change shape or disappear when the bifurcation is crossed, then this bifurcation will 
not affect learning. 

Crossing boundaries between basins of attraction is a \emph{local} event, and it is 
\emph{sufficient} for the gradients to explode. If we assume that 
crossing into an emerging attractor or from a disappearing one (due to a bifurcation) qualifies 
as crossing some boundary between attractors, that we can formulate a \emph{sufficient} 
condition for gradients to explode which encapsulates the observations made in \citet{Doya93}, 
extending them to also normal crossing of boundaries between different basins of attractions. 
Note how in the figure, there are only two values of $b$ with a bifurcation, but a whole range
of values for which there can be a boundary crossing.

Another limitation of previous analysis is that they only consider autonomous systems
and assume the observations hold for input-driven models. In \cite{Bengio-trnn94} input 
is dealt with by assuming it is bounded noise. The downside of this approach is that 
it limits how one can reason about the input. In practice, the input is supposed to drive 
the dynamical system, being able to leave the model in some attractor state, or kick it out
of the basin of attraction when certain triggering patterns present themselves. 

\begin{figure}[ht]
\vspace*{-2mm}
\begin{center}
\centerline{\includegraphics[width=.7\columnwidth]{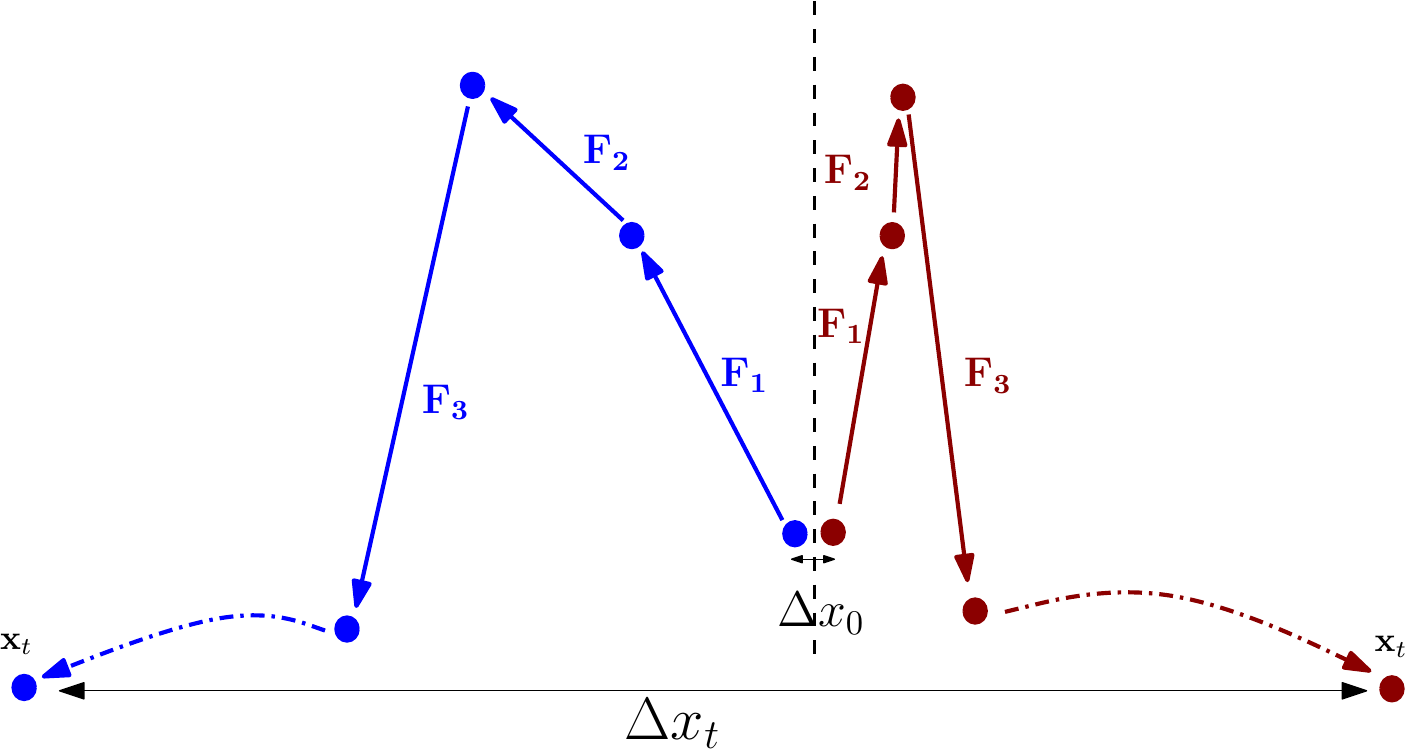}}
\vspace*{-2mm}
  \caption{ This diagram illustrates how the change in $\state_t$, $\Delta \state_t$,
      can be large for a small $\Delta \state_0$. 
      The blue vs red (left vs right) trajectories are generated by the same maps $F_1, F_2, \ldots$ 
      for two different initial states.} 
  \label{fig:input_driven1}
\end{center}
\vspace*{-5mm}
\end{figure}

We propose to extend our analysis to input driven models by folding
the input into the map. We consider the family of 
maps $F_t$, where we apply a different $F_t$ at each step. Intuitively, 
for the gradients to explode 
we require the same behaviour as before, where (at least in some direction) the 
maps $F_1, .., F_t$ agree and change direction.
Fig. \ref{fig:input_driven1} describes this behaviour. 

\begin{figure}[ht]
\vspace*{-1mm}
\begin{center}
\centerline{\includegraphics[width=.7\columnwidth]{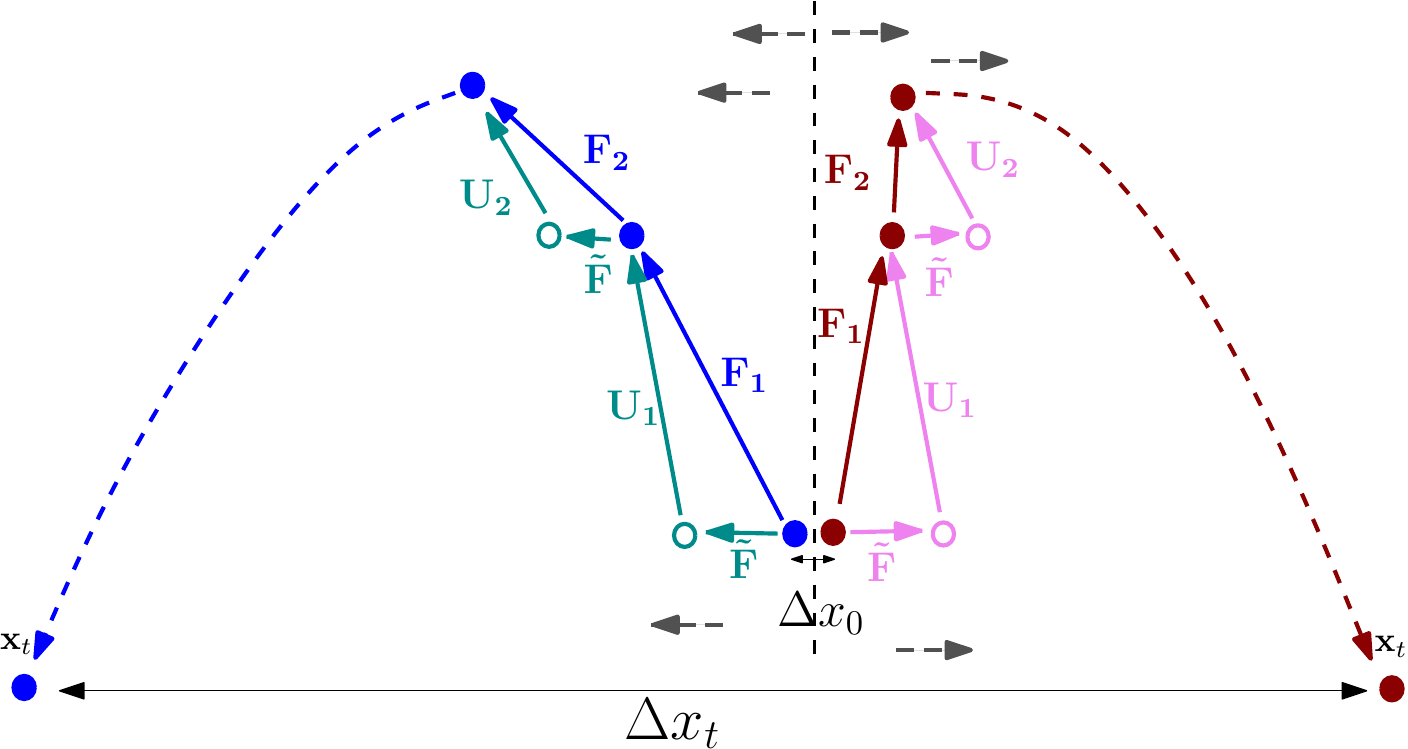}}
\vspace*{-2mm}
  \caption{ Illustrates how one can break apart the maps $F_1, ..F_t$ into a constant 
  map $\tilde{F}$ and the maps $U_1, .., U_t$.  The dotted vertical line represents the boundary 
  between basins of attraction, and the straight dashed arrow the direction of the map $\tilde{F}$ 
  on each side of the boundary. This diagram is an extension of Fig. \ref{fig:input_driven1}.} 
  \label{fig:input_driven2}
\end{center}
\vspace*{-5mm}
\end{figure}

For the specific parametrization provided by equation \eqref{eq:standard_rnn} we can 
take the analogy one step further by decomposing the maps $F_t$ into a fixed map $\tilde{F}$ 
and a time-varying one $U_t$. $F(\state) = \mathbf{W}_{rec}\sigma(\state) +\mathbf{b}$ 
corresponds to an input-less recurrent network, while $U_t(\state)= \state + \mathbf{W}_{in}\inp_t$
describes the effect of the input. This is depicted in in Fig. \ref{fig:input_driven2}.
Since $U_t$ changes with time, it can not be analyzed 
using standard dynamical systems tools, but $\tilde{F}$ can. 
This means that when a boundary between basins of attractions is crossed for 
$\tilde{F}$, the state will move towards a different attractor, which for large $t$ 
could lead (unless the input maps $U_t$ are opposing this) to a large discrepancy in $\state_t$.
Therefore studying 
the asymptotic behaviour of $\tilde{F}$ can provide useful information about where such events are likely
to happen.

One interesting observation from the dynamical systems perspective with respect to vanishing gradients 
is the following. If the factors $\frac{\partial \state_t}{\partial \state_k}$ 
go to zero (for $t-k$ large), it means that $\state_t$ does not depend on $\state_k$ (if we change
$\state_k$ by some $\Delta$, $\state_t$ stays the same). This translates into the model at 
$\state_t$ being close to convergence towards some attractor (which it would reach from anywhere 
in the neighbourhood of $\state_k$). 

\vspace*{-2mm}
\subsection{The geometrical interpretation}
\label{sec:geometric}
\vspace*{-1mm}

Let us consider a simple one hidden unit model
(equation \eqref{eq:simple_model}) where we provide an initial state $x_{0}$ and train the model to have 
a specific target value after 50 steps. Note that for simplicity we assume no input.

\begin{equation}
\vspace*{-1mm}
\label{eq:simple_model}
x_t = w \sigma(x_{t-1}) + b
\vspace*{-.5mm}
\end{equation}

\begin{figure}[ht]
\vspace*{-2mm}
\begin{center}
\centerline{\includegraphics[width=.8\columnwidth]{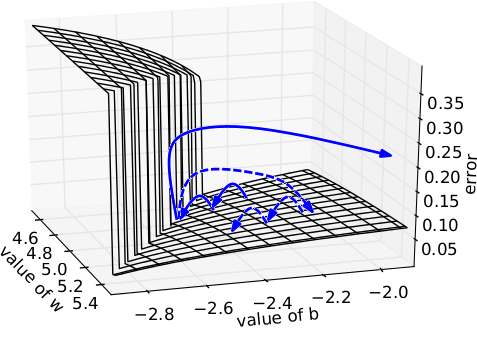}}
\vspace*{-1mm}
  \caption{We plot the error surface of a single hidden unit recurrent network, 
       highlighting the existence of high curvature walls. 
      The solid lines depicts standard trajectories that gradient descent 
      might follow. Using dashed arrow the diagram shows what would happen if 
      the gradients is rescaled to a fixed size when its norm is above a threshold.
  } 
  \label{fig:curvature}
\end{center}
\vspace*{-5mm}
\end{figure}

Fig. \ref{fig:curvature} shows the error surface $\loss_{50} = (\sigma(x_{50}) - 0.7)^2$, where 
 $x_0 = .5$ and $\sigma$ to be the sigmoid function. 

We can more easily analyze the behavior of this model by further simplifying it to be
linear ($\sigma$ then being the identity function), with
$b=0$. $x_t = x_0 w^t$ from which it follows that 
$\frac{\partial x_t}{\partial w} = tx_0w^{t-1}$ and $\frac{\partial^2 x_t}{\partial w^2} = t(t-1)x_0 w^{t-2}$,
implying that when the first derivative explodes, so does the second derivative. 

In the general case, when the gradients explode they do so along some directions $\mathbf{v}$. 
This says that there exists, in such situations, a vector $\mathbf{v}$ such that 
$\frac{\partial \loss_t}{\partial \theta} \mathbf{v} \geq C\alpha^t$, where $C, \alpha \in \mathrm{R}$ and 
$\alpha > 1$. For the linear case ($\sigma$ is the identity function), $\mathbf{v}$ is the eigenvector 
corresponding to the largest eigenvalue of $\mathbf{W}_{rec}$. If this bound is tight, we hypothesize that in general {\em when gradients 
explode so does the curvature along $\mathbf{v}$, leading to a wall in the 
error surface}, like the one seen in Fig. \ref{fig:curvature}. 

If this holds, then it gives us a simple solution
to the exploding gradients problem depicted in Fig. \ref{fig:curvature}. 

If both the gradient and the leading eigenvector of the curvature are aligned 
with the exploding direction $\mathbf{v}$, it follows that the error surface has a steep wall perpendicular 
to $\mathbf{v}$ (and consequently to the gradient). 
This means that when stochastic gradient descent (SGD) reaches the wall and does a gradient descent step, 
it will be forced to jump across the valley moving perpendicular to the steep walls, possibly 
leaving the valley and  disrupting the learning process. 

The dashed arrows in Fig.~\ref{fig:curvature} correspond to {\em ignoring the norm of this large step}, ensuring that the model 
stays close to the wall. The key insight is that all the steps taken when the gradient explodes
are aligned with $\mathbf{v}$ and ignore other descent direction (i.e. the model moves perpendicular 
to the wall). At the wall, a small-norm step in the direction of the gradient therefore merely pushes
us back inside the smoother low-curvature region besides the wall, whereas a regular gradient step would bring us very far, thus slowing or preventing further training. 
Instead, with a bounded step, we get back in that smooth region near the wall
where SGD is free to explore  other descent directions. 

The important addition in this scenario to the classical high curvature valley, is 
that we assume that the valley is wide, as we have a large region around the wall where 
if we land we can rely on first order methods to move towards the local minima. 
This is why just clipping the gradient might be sufficient,
not requiring the use a second order method. 
Note that this algorithm should work even when the rate of growth of the gradient 
is not the same as the one of the curvature (a case for which a second order method would fail as 
the ratio between the gradient and curvature could still explode).
%

Our hypothesis could also help to understand the recent success of the Hessian-Free approach compared 
to other second order methods. There are two key differences between Hessian-Free and most other second-order algorithms. 
First, it uses the full Hessian matrix and hence can deal with exploding directions that are
not necessarily axis-aligned. Second, it computes a new estimate of the Hessian matrix 
before each update step and can take into account abrupt changes in curvature (such as the ones suggested by our hypothesis) while 
most other approaches use a smoothness assumption, i.e.,
averaging 2nd order signals over many steps.

\vspace*{-3mm}
\section{Dealing with the exploding and vanishing gradient}
\label{sec:solutions}
\vspace*{-1mm}
\subsection{Previous solutions}
\label{sec:prev_sol}
\vspace*{-1mm}

Using an L1 or L2  penalty on the recurrent weights can help with exploding gradients.
Given that the parameters initialized with small values, the spectral radius of $\mathbf{W}_{rec}$
is probably smaller than 1, from which it follows that the gradient can not explode (see 
necessary condition found in section \ref{sec:mechanics}).
The regularization term can ensure that during training the spectral radius 
never exceeds 1. 
This approach limits the model to a simple 
regime (with a single point attractor at the origin), where any information inserted in the model has to 
die out exponentially fast in time. In such a regime we can not train a generator network, nor can we 
exhibit long term memory traces.

\citet{Doya93} proposes to pre-program the model (to initialize the model 
in the right regime) or to use \emph{teacher forcing}. The first proposal assumes that 
if the model  exhibits from the beginning the same kind of asymptotic behaviour as the one required 
by the target, then there is no need to cross a bifurcation boundary. The downside is that one can 
not always know the required asymptotic behaviour, and, even if such information 
is known, it is not trivial to initialize a model in this specific regime. We should also note that such 
initialization does not prevent crossing the boundary between basins of attraction, which, as shown, 
could happen even though no bifurcation boundary is crossed. 

Teacher forcing is a more interesting, yet a not very well understood solution. It can be seen 
as a way of initializing the model in the right regime and the right region of space. It has been 
shown that in practice it can reduce the chance that gradients explode, and even allow
training generator models or models that work with unbounded amounts of memory\citep{Pascanu11,DoyaY91}.
One important downside is that it requires a target 
to be defined at every time step. 

In \citet{hochreiter1997long,Graves2009HandwritingRecognition} a solution is proposed for the vanishing gradients problem, where the structure of 
the model is changed. Specifically it introduces a special set of units called LSTM units which are linear 
and have a recurrent connection to itself which is fixed to 1. The flow of information into the unit and from 
the unit is guarded by an input and output gates (their behaviour is learned). There are several variations 
of this basic structure. This solution does not address explicitly the exploding gradients problem. 

\citet{ICML2011Ilya} use the Hessian-Free optimizer in conjunction 
with \emph{structural damping}, a specific damping 
strategy of the Hessian. This approach seems to deal very well with the 
vanishing gradient, though more detailed analysis is still missing. Presumably
this method works because in high dimensional spaces there is a high 
probability for long term components to be orthogonal to short term ones. This would allow
the Hessian to rescale these components independently. In practice, one can not guarantee that this 
property holds. As discussed in section \ref{sec:geometric},
this method is able to deal with the exploding gradient as well. 
Structural damping is an enhancement that forces the change in the state to be small, 
when the parameter changes by some small value $\Delta \theta$. This asks for 
the Jacobian matrices $\frac{\partial \state_t}{\partial \theta}$ to have small norm,
hence further helping with the exploding gradients problem. The fact that it helps 
when training recurrent neural models on long sequences suggests 
that while the curvature might explode at the same time with the gradient, it might not 
grow at the same rate and hence not be sufficient to deal with the exploding gradient. 

Echo State Networks \citep{Lukosevicius2009Reservoir} avoid the exploding and vanishing gradients problem by not learning 
the recurrent and input weights. They are sampled from hand crafted distributions. 
Because usually the largest eigenvalue of the recurrent weight is, by construction,
smaller than 1, information fed in to the model has to die out exponentially fast. This means
that these models can not easily deal with long term dependencies, even though the reason is slightly different 
from the vanishing gradients problem. An extension to the classical 
model is represented by leaky integration units \citep{JaegerLPS07},
 where 

$\state_k= \alpha \state_{k-1} + (1-\alpha) \sigma(\mathbf{W}_{rec}\state_{k-1} + \mathbf{W}_{in} \inp_k + \mathbf{b})$. 

While these units can be used to solve the standard benchmark proposed by \citet{hochreiter1997long} for learning long 
term dependencies (see \cite{jaeger12}), they are more suitable to deal with low frequency information as they act as a
low pass filter. 
Because most of the 
weights are randomly sampled, is not clear what size of models one would need  to solve complex  
real world tasks. 

We would make a final note about the approach proposed by Tomas Mikolov in his PhD thesis~\citep{tomas_phd}(and implicitly 
used in the state of the art results on language modelling \citep{Mikolov-Interspeech-2011}).
It involves clipping the gradient's temporal components element-wise (clipping an entry when it exceeds in absolute value a fixed threshold).
Clipping has 
been shown to do well in practice and it forms the backbone of our approach.

\vspace*{-2mm}
\subsection{Scaling down the gradients}
\label{sec:exploding_sol}
\vspace*{-1mm}

As suggested in section \ref{sec:geometric}, one simple mechanism to deal with a sudden increase
in the norm of the gradients is to rescale them whenever they go over a threshold
(see algorithm \ref{algo:0}). 

\vspace*{-2mm}
\begin{algorithm}
\vspace*{-.5mm}
\caption{{Pseudo-code for norm clipping the gradients whenever they explode}}
\label{algo:0}
\begin{algorithmic}
\STATE $\hat{\mathbf{g}} \leftarrow \frac{\partial \loss}{\partial \theta}$
\IF{ $\|\hat{\mathbf{g}}\| \geq threshold$}
    \STATE $\hat{\mathbf{g}} \leftarrow \frac{threshold}{\|\hat{\mathbf{g}} \|} \hat{\mathbf{g}}$
\ENDIF
\end{algorithmic}
\end{algorithm}
\vspace*{-2mm}

This algorithm is very similar to the one proposed by Tomas Mikolov and 
we only diverged from the original proposal in an attempt to provide a better 
theoretical foundation (ensuring that we always move in a descent direction
with respect to the current mini-batch), 
though in practice both variants behave similarly. 

The proposed clipping is simple to implement and computationally efficient, but 
it does however introduce an additional
hyper-parameter, namely the threshold.
One good heuristic for setting this threshold is to look at statistics on the average norm 
over a sufficiently large number of updates. In our experiments we have noticed 
that for a given task and model size, training is not very sensitive to this hyper-parameter and the algorithm 
behaves well even for rather small thresholds.

The algorithm can also be thought of as adapting the learning rate based 
on the norm of the gradient. Compared to other 
learning rate adaptation strategies, which focus on improving convergence 
by collecting statistics on the gradient (as for example in \citet{DuchiHS11}, 
or \citet{Moreira-95} for an overview),
we rely on the {\em instantaneous} gradient.
This means that we can handle very abrupt changes in norm, while the other methods 
would not be able to do so. 

\vspace*{-2mm}
\subsection{Vanishing gradient regularization}
\label{sec:vanish}
\vspace*{-1mm}

We opt to address the vanishing gradients problem using a regularization term 
that represents a preference for parameter values such that
back-propagated gradients neither increase or decrease too much in magnitude.
Our intuition is that  increasing the norm of 
$\frac{\partial \state_t}{\partial \state_k}$ means the error at time $t$ is more sensitive to all inputs
$\inp_t, .., \inp_k$ ($\frac{\partial \state_t}{\partial \state_k}$ is a factor in $\frac{\partial \loss_t}{\partial \inp_k}$). 
In practice some of these inputs will be irrelevant for the prediction at time $t$ and will 
behave like noise that the network needs to learn to ignore. The network can not learn to ignore these irrelevant 
inputs unless there is an error signal. These two issues can not be solved in parallel, and it seems natural to expect
that we need to force the network to increase the norm of $\frac{\partial \state_t}{\partial \state_k}$ at the expense 
of larger errors (caused by the irrelevant input entries) and \emph{then} wait for it to learn to ignore these irrelevant 
input entries. This suggest that moving towards increasing the 
norm of $\frac{\partial \state_t}{\partial \state_k}$ can not be always done while following a descent direction of 
the error $\loss$ (which is, for e.g.,  what a second order method would try to do), and therefore we need to enforce 
it via a regularization term. 

The regularizer we propose below
prefers solutions for which the error signal preserves 
norm as it travels back in time:
\begin{equation}
\vspace*{-4mm}
\label{eq:reg_term}
\Omega = \sum_k \Omega_k = \sum_k \left( 
\frac{ \left\| \frac{\partial \loss}{\partial \state_{k+1}} \frac{\partial \state_{k+1}}{\partial \state_{k}}\right\|}{ \left\| \frac{\partial \loss}{\partial \state_{k+1}} \right\|}
- 1 \right)^2
\vspace*{-.5mm}
\end{equation}

In order to be computationally efficient, we only use the ``immediate'' partial derivative of $\Omega$ with respect to $\mathbf{W}_{rec}$ 
(we consider that $\state_k$ and $\frac{\partial \loss}{\partial \state_{k+1}}$ as being constant with respect to $\mathbf{W}_{rec}$ when 
computing the derivative of $\Omega_k$), as depicted in equation \eqref{eq:dir_deriv}. Note we use the parametrization 
of equation \eqref{eq:standard_rnn}. This can be done efficiently because 
we get the values of $\frac{\partial \loss}{\partial \state_k}$ from BPTT. We use Theano to compute these gradients \citep{bergstra+al:2010-scipy,Bastien-Theano-2012}. 
\begin{equation}
\vspace*{-2mm}
\label{eq:dir_deriv}
\begin{array}{lll}
\frac{\partial^+ \Omega}{\partial \mathbf{W}_{rec}} & = &  \sum_k \frac{\partial^+ \Omega_k}{\partial \mathbf{W}_{rec}} \\
& = & \sum_k \frac{\partial^+ 
  \left(
  \frac{ \left\| \frac{\partial \loss}{\partial \state_{k+1}} \mathbf{W}_{rec}^T {diag}(\sigma'(\state_k)) \right\|}{\left\| \frac{\partial \loss}{\partial \state_{k+1}} \right\|} -1 \right)^2}
  {\partial \mathbf{W}_{rec}}
\end{array}
\vspace*{-.5mm}
\end{equation}

Note that our regularization term only forces the Jacobian matrices $\frac{\partial \state_{k+1}}{\partial \state_k}$ to 
preserve norm in the relevant direction of the error $\frac{\partial \loss}{\partial \state_{k+1}}$, not 
for any direction (i.e. we do not enforce that all eigenvalues are close to 1). The second observation is that 
we are using a soft constraint, therefore we are not ensured the norm of the error signal is preserved. If it 
happens that these Jacobian matrices are such that the norm explodes (as $t-k$ increases), then this could lead 
to the exploding gradients problem and we need to deal with it for example as described in section \ref{sec:exploding_sol}.
This can be seen from  the dynamical systems perspective as well: preventing vanishing gradients 
implies that 
we are pushing the model such that it is further away from the attractor (such that it does not converge to it, case in which
the gradients vanish) and closer to boundaries between basins of attractions, making it more probable for the gradients 
to explode. 

\vspace*{-2mm}
\section{Experiments and Results}
\label{sec:experiments}
\vspace*{-1mm}
\subsection{Pathological synthetic problems}
\vspace*{-1mm}

As done in \citet{Martens+Sutskever-ICML2011}, we address the pathological problems proposed 
by \citet{hochreiter1997long} that require learning long term correlations. We refer 
the reader to this original paper for a detailed description of the tasks and to the 
supplementary materials for the complete description of the experimental setup. 

\vspace*{-1mm}
\subsubsection{The Temporal Order problem}
\vspace*{-1mm}

We consider the temporal order problem as the prototypical pathological problem, 
extending our results to the other proposed tasks afterwards. The input 
is a long stream of discrete symbols. 
At two points in time 
(in the beginning and middle of the sequence) a symbol within $\{A,B\}$ is emitted. 
The task consists in classifying the order (either $AA, AB, BA, BB$) at the end of the sequence. 

\begin{figure}[ht]
\begin{center}
\vspace*{-3mm}
\centerline{\includegraphics[width=1.\columnwidth]{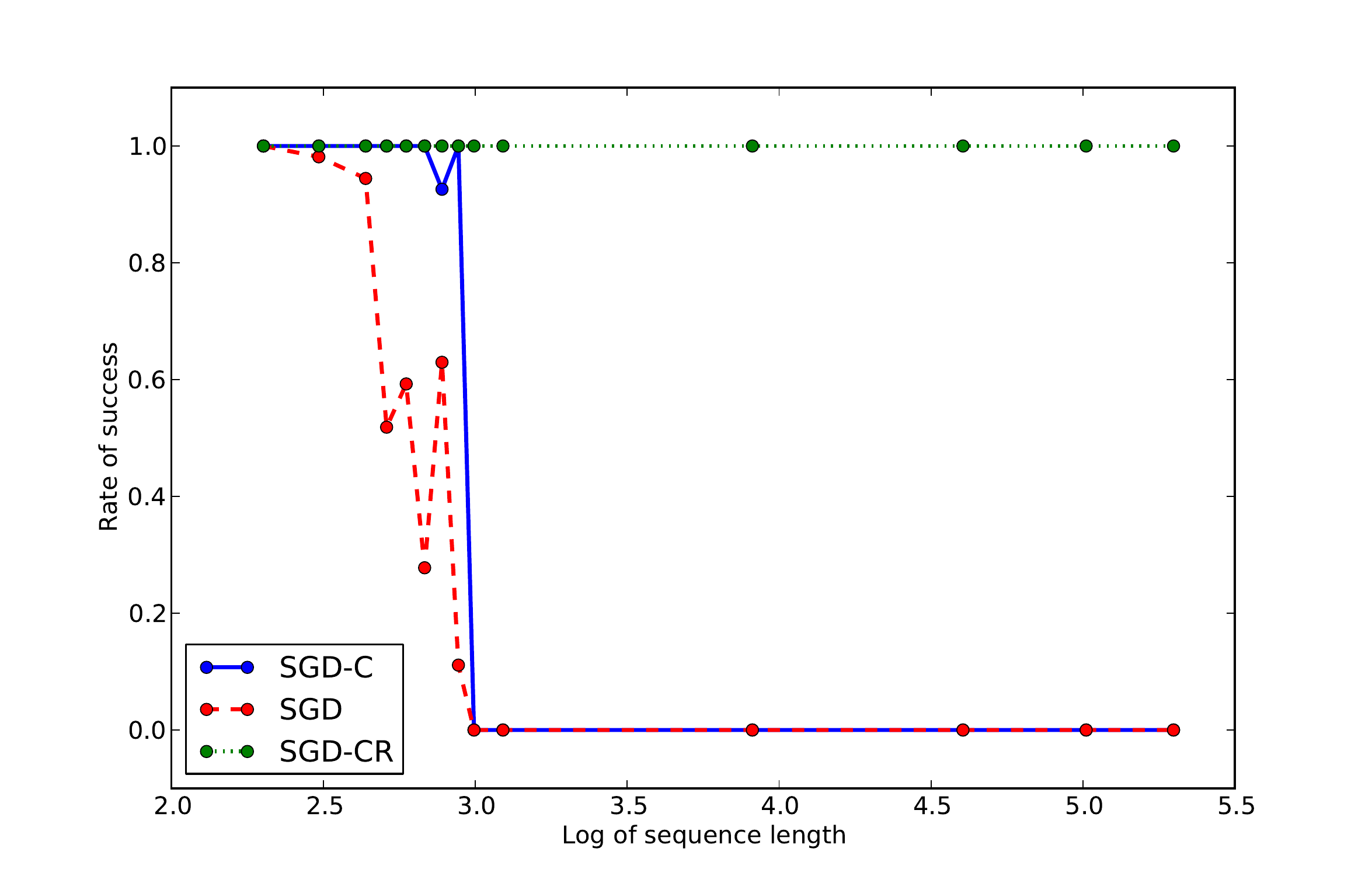}}
\vspace*{-5mm}
  \caption{Rate of success for solving the temporal order problem versus log of sequence length. See text.} 
  \label{fig:tempOrder_results}
\end{center}
\vspace*{-6mm}
\end{figure}

Fig. \ref{fig:tempOrder_results} shows the success rate of standard SGD, SGD-C (SGD enhanced with 
out clipping strategy) and SGD-CR (SGD with the clipping strategy and the regularization term). 
Note that for sequences longer than 20, the vanishing gradients problem ensures that neither 
SGD nor SGD-C algorithms can solve the task. The $x$-axis is on log scale. 

This task provides empirical evidence that exploding gradients are linked with tasks that require 
long memory traces. We know that initially the model operates in the one-attractor regime (i.e. 
$\lambda_{1}~<~1$),
in which the amount of memory is controlled by $\lambda_{1}$. 
More memory means larger spectral radius, and, when this value crosses a certain threshold the model enters rich regimes 
where gradients are likely to explode. We see in Fig. \ref{fig:tempOrder_results} that as long as the 
vanishing gradient problem does not become an issue, addressing the exploding gradients problem ensures a better success 
rate.

When combining clipping as well as the regularization term proposed in section \ref{sec:vanish}, we call this algorithm SGD-CR. SGD-CR solved the task with a success rate of 100\% for sequences up to 200 steps (the maximal length used in
\citet{Martens+Sutskever-ICML2011}). Furthermore, we can train a single model to deal with any sequence of length 50 up 
to 200 (by providing sequences of different lengths for different SGD steps). Interestingly enough, \textbf{the trained model can 
generalize to new sequences that can be twice as long as the ones seen during training}.

\vspace*{-2mm}
\subsubsection{Other pathological tasks}
\vspace*{-1mm}

SGD-CR was also able to solve (100\% success on the lengths listed below, for all but one task)
other pathological problems proposed in \citet{hochreiter1997long}, namely the \emph{addition}
problem, the \emph{multiplication} problem, the \emph{3-bit  temporal order problem}, the \emph{random permutation} problem and
the \emph{noiseless memorization} problem in two variants (when the pattern needed to be memorized is 5 bits in length and when it contains 
over 20 bits of information; see \citet{Martens+Sutskever-ICML2011}).
For the first 4 problems we used a single model for lengths up to 200, while for the \emph{noiseless memorization} we used a different 
model for each sequence length (50, 100, 150 and 200). The hardest problems for which only one trail out of 8 succeeded was the random
permutation problem. In all cases, we observe successful generalization to sequences
longer than the training sequences.
In most cases, these results outperforms \citet{Martens+Sutskever-ICML2011} in terms 
of success rate, they deal with longer sequences than in \citet{hochreiter1997long} and compared to 
\cite{jaeger12} they generalize to longer sequences.

\vspace*{-2mm}
\subsection{Natural problems}
\vspace*{-1mm}

We address the task of polyphonic music prediction, using the datasets 
Piano-midi.de, Nottingham and MuseData described in \citet{Pascal+all-ICML2012} 
and language modelling at the character level 
on the Penn Treebank dataset \citep{TomasIlya}. We also explore a modified version 
of the task, where we ask the model to predict the 5th character in the future (instead 
of the next). Our assumption is that to solve this modified task long term correlations 
are more important than short term ones, and hence our regularization term should be more 
helpful. 

The training and test scores reported in Table \ref{tabl:music_scores} are 
average negative log likelihood per time step. We fixed hyper-parameters across
the three runs, except for the regularization factor and 
clipping cutoff threshold. 
{\bf SGD-CR provides a statistically significant
improvement on the state-of-the-art for RNNs on all the polyphonic music prediction tasks}
{\em except} for MuseData on which we get exactly the same performance as the
state-of-the-art \citep{YoshuaNP-2012}, which uses a different architecture.
Table \ref{tabl:text} contains the results on language modelling (in bits per letter).

\begin{table}[t]
\vspace*{-2mm}
    \caption{Results on polyphonic music prediction in negative log likelihood per time step. Lower is better.}
    \label{tabl:music_scores}
\vspace*{-3mm}
    \begin{center}
        \begin{small}
            \begin{sc}
                \begin{tabular}{m{1.8cm}|m{.8cm}|m{1cm}|m{1.2cm}|m{1.4cm}|}
                    \hline
                    Data set & Data fold & SGD & SGD+C & SGD+CR \\
                    \hline
                    \hline
                    Piano-  & train & 6.87 & \bf 6.81 & 7.01 \\ \cline{2-5}
                    midi.de & test  & 7.56 & 7.53 & \bf 7.46 \\ 
                    \hline \hline
                    Nottingham & train & 3.67 & \bf 3.21 & 3.24 \\ \cline{2-5}
                               & test  & 3.80 & 3.48 & \bf 3.46 \\ \hline \hline
                    MuseData   & train & 8.25 & 6.54 & \bf 6.51 \\ \cline{2-5}
                               & test  & 7.11 & 7.00 & \bf 6.99 \\
                    \hline
                \end{tabular}
            \end{sc}
        \end{small}
    \end{center}
    \vskip -0.1in
\vspace*{-5mm}
\end{table}

\begin{table}[t]
\vspace*{-1mm}
    \caption{Results on the next character prediction task in entropy (bits/character)}
    \label{tabl:text}
\vspace*{-1mm}
    \begin{center}
        \begin{small}
            \begin{sc}
                \begin{tabular}{m{1.6cm}|m{.8cm}|m{1cm}|m{1.2cm}|m{1.4cm}|}
                    \hline
                    Data set & Data fold & SGD & SGD+C& SGD+CR \\
                    \hline
                    \hline
                    1 step & train & 1.46 & \bf 1.34 & 1.36 \\ \cline{2-5}
                             & test  & 1.50 & 1.42 & \bf 1.41 \\ 
                    \hline \hline
                     5 steps & train & N/A & 3.76 & \bf 3.70  \\ \cline{2-5}
                             & test  & N/A & 3.89 & \bf 3.74 \\
                    \hline
                 \end{tabular}
             \end{sc}
         \end{small}
     \end{center}
\vspace*{-8mm}
\end{table}

These results 
suggest that clipping the gradients solves an
optimization issue and does not act as a regularizer, 
as both the training and test error improve in general.
Results on Penn Treebank reach the state of the art 
achieved by \citet{TomasIlya}, who used 
a different clipping algorithm similar to ours,
thus providing evidence that both behave similarly. 
The regularized model performs as well as the Hessian-Free trained model.

By employing the proposed regularization term we are able to improve 
test error even on tasks that are not dominated by long term 
contributions.

\vspace*{-3mm}
\section{Summary and Conclusions}
\label{sec:conclusion}
\vspace*{-1mm}

We provided different 
perspectives through which one can gain more insight into the
\emph{exploding and vanishing gradients} issue. To deal with the
exploding gradients problem, we propose a solution that involves clipping the norm of the exploded gradients when it
is too large. The algorithm is motivated 
by the assumption that when gradients explode, the curvature and higher order derivatives explode as well,
and we are faced with a specific pattern in the error surface, namely a valley 
with a single steep wall.
In order to deal with the vanishing gradient problem we use a regularization term that forces the error signal 
not to vanish as it travels back in time. This regularization term forces the Jacobian matrices 
$\frac{\partial \state_i}{\partial \state_{i-1}}$ to preserve norm only in relevant directions.
In practice we show that these solutions improve performance 
on both the pathological synthetic datasets considered as well as on polyphonic music prediction and language modelling. 

\vspace*{-3mm}

\subsection*{Acknowledgements}

We would like to thank the Theano 
development team as well (particularly to Frederic Bastien, Pascal Lamblin and James Bergstra) for their help. 

We acknowledge NSERC, FQRNT, CIFAR, RQCHP and Compute Canada for the resources they provided.

\bibliographystyle{natbib}
\bibliography{arxiv}
\newpage
\section*{Analytical analysis of the exploding and vanishing gradients problem}
\label{sec:analysis}

\begin{equation}
    \label{eq:standard_rnn}
    \state_{t} = \mathbf{W}_{rec} \sigma(\state_{t-1}) + \mathbf{W}_{in} \inp_{t} + \mathbf{b}
\end{equation}
Let us consider the term 
$\mathbf{g}_k^T=\frac{\partial \loss_t}{\partial \state_t}\frac{\partial \state_t}{\partial \state_k}
\frac{\partial^+ \state_k}{\partial \theta}$ for the linear version of the 
parametrization in equation \eqref{eq:standard_rnn} (i.e. set $\sigma$ to the identity function) and 
assume $t$ goes to infinity and $l=t-k$. We have that:
\begin{equation}
\label{eq:prod_wk}
\frac{\partial \state_t}{\partial \state_k} = \left(\mathbf{W}_{rec}^{T}\right)^{l}
\end{equation}
By employing a generic  \emph{power iteration method} based proof we can show that,
given certain conditions, 
$\frac{\partial \loss_t}{\partial \state_t}\left(\mathbf{W}_{rec}^{T}\right)^{l}$ 
grows exponentially. 
\begin{proof}
Let $\mathbf{W}_{rec}$ have the eigenvalues $\lambda_1, ..,\lambda_n$ with 
$|\lambda_1| > |\lambda_2| > .. > |\lambda_n|$ and
the corresponding eigenvectors $\mathbf{q}_1, \mathbf{q}_2, .., \mathbf{q}_n$ 
which form a vector basis. We can now write the row vector $\frac{\partial \loss_t}{\partial \state_t}$
into this basis:

$\frac{\partial \loss_t}{\partial \state_t} = \sum_{i=1}^N c_i \mathbf{q}_i^T$

If $j$ is such that $c_j \neq 0$ and any $j'<j, c_{j'}=0$, using the fact that 
$\mathbf{q}_i^T \left(\mathbf{W}_{rec}^T\right)^{l} = \lambda_i^{l} \mathbf{q}_i^T$ we have that 
\begin{equation}
\label{eq:approx_comp}
\frac{\partial \loss_t}{\partial \state_t}\frac{\partial \state_t}{\partial \state_k} = 
c_j \lambda_j^{l} \mathbf{q}_j^T + \lambda_j^{l}\sum_{i={j+1}}^n 
c_i \frac{\lambda_i^l}{\lambda_j^l}\mathbf{q}_i^T
\approx c_j \lambda_j^{l} \mathbf{q}_j^T
\end{equation}

We used the fact that $\left|\slantfrac{\lambda_i}{\lambda_j}\right| <1$ for $i>j$, which means that $\lim_{l \to \infty} 
\left|\slantfrac{\lambda_i}{\lambda_j}\right|^l = 0$. 
If $|\lambda_j| > 1$, it follows that $\frac{\partial \state_t}{\partial \state_k}$ grows exponentially 
fast with $l$, and it does so along the direction $\mathbf{q}_j$.
\qed
\end{proof}
The proof assumes $\mathbf{W}_{rec}$ is diagonalizable for simplicity, 
though using the Jordan normal form of $\mathbf{W}_{rec}$ one can extend this proof
by considering not just the eigenvector of largest eigenvalue but the whole subspace
spanned by the eigenvectors sharing the same (largest) eigenvalue.

This result provides a necessary condition for gradients to grow, namely that the spectral 
radius (the absolute value of the largest eigenvalue) 
of $\mathbf{W}_{rec}$ must be larger than 1. 

If $\mathbf{q}_j$ is not in the null space of $\frac{\partial^+ 
\state_k}{\partial \theta}$ the entire temporal component grows exponentially with $l$. 
This approach extends easily to the entire gradient. If we re-write it in terms of the eigen-decomposition
of $\mathbf{W}$, we get:
\begin{equation}
\frac{\partial \loss_t}{\partial \theta} = \sum_{j=1}^n \left( \sum_{i=k}^t c_j \lambda_j^{t-k} \mathbf{q}_j^T \frac{\partial^{+} \state_k}{\partial \theta}\right)
\end{equation}
We can now pick $j$ and $k$ such that $c_j\mathbf{q}_j^T \frac{\partial^+ \state_k}{\partial \theta}$ does not have 0 norm, while maximizing 
$|\lambda_j|$. If for the chosen $j$ it holds that  $|\lambda_j| > 1$ then $\lambda_j^{t-k} c_j \mathbf{q}_j^T \frac{\partial^+ \state_k}{\partial \theta}$ 
will dominate the sum and because this term grows exponentially fast to infinity with $t$, the same will happen to the sum.

\section*{Experimental setup}
\label{sec:setup}
Note that all hyper-parameters where selected based on their performance on a validation set using  a grid search. 

\subsection*{The pathological synthetic tasks}
Similar success criteria is used in all of the tasks below (borrowed from \citet{Martens+Sutskever-ICML2011}),
namely that the model should make no more than 1\% error on a batch of 10000 test samples. 
In all cases, discrete symbols are depicted by a one-hot encoding, and in case of regression 
a prediction for a given sequence is considered as a success if the error is less than 0.04.

\subsubsection*{Addition problem}
The input consists of a sequence of random numbers, 
where two random positions (one in the beginning and one in the middle 
of the sequence) are marked. The model needs to predict the 
sum of the two random numbers after the entire sequence was seen. 
For each generated sequence we sample the length $T'$ from $[T, \frac{11}{10}T]$, though 
for clarity we refer to $T$ as the length of the sequence in the paper. 
The first position is sampled from $[1, \frac{T'}{10}]$, while the 
second position is sampled from $[\frac{T'}{10}, \frac{T'}{2}]$. These positions $i,j$ are  marked
in a different input channel that is 0 everywhere except for the 
two sampled positions when it is 1.  The model needs to predict the sum of the 
random numbers found at the sampled positions $i,j$ divided by 2. 

To address this problem we use a 50 hidden units model, with a tanh activation function.
The learning rate is set to .01 and the factor $\alpha$ in front of the regularization term 
is 0.5. We use clipping with a cut-off threshold of 6 on the norm of the gradients.
The weights are initialized from a normal distribution with mean 0 and standard derivation $.1$. 

The model is trained on sequences of varying length $T$ between 50 and 200.
We manage to get a success rate of 100\%   
at solving this task, which outperforms 
the results presented in \citet{Martens+Sutskever-ICML2011} (using Hessian Free),
where we see a decline in success rate as the length of the sequence gets closer to 
200 steps. \citet{hochreiter1997long} only considers sequences up to 100 steps.
\citet{jaeger12} also addresses this task with 100\% success rate, though the solution
does not seem to generalize well as it relies on very large output weights, which 
for ESNs are usually a sign of instability. 
We use a single model to deal with 
all lengths of sequences (50, 100, 150 200), and the trained model generalizes to new sequences 
that can be up 400 steps (while the error is still under 1\%).

\subsubsection*{Multiplication problem}

This task is similar to the problem above, just that the predicted value is the product of the 
random numbers instead of the sum. We used the same hyper-parameters as for the previous 
case, and obtained very similar results.

\subsubsection*{Temporal order problem}
For the temporal order the length of the sequence is fixed to $T$, We have a fixed 
set of two symbols $\{A, B\}$ and 4 distractor symbols $\{c, d, e, f\}$. 
The 
sequence entries are uniformly sampled from the distractor symbols everywhere except 
at two random positions, the first position sampled from $[\frac{T}{10}, \frac{2T}{10}]$, 
while the second from $[\frac{4T}{10}, \frac{5T}{10}]$. The task is to predict the 
order in which the non-distractor symbols were provided, i.e. either $\{AA, AB, BA, BB\}$. 

We use a 50 hidden units model, with a learning rate of .001 and $\alpha$, the regularization coefficient, 
set to 2. 
The cut-off threshold for clipping the norm of the gradient is left to 6. 
As for the other two task we have a 100\% success rate at training a single model 
to deal with sequences between 50 to 200 steps. This outperforms the previous state
of the art because of the success rate, but also the single model generalizes to 
longer sequences (up to 400 steps). 

\subsubsection*{3-bit temporal order problem}
Similar to the previous one, except that we have 3 random positions, first 
sampled from $[\frac{T}{10}, \frac{2T}{10}]$, second from $[\frac{3T}{10}, \frac{4T}{10}]$ 
and last from $[\frac{6T}{10}, \frac{7T}{10}]$. 

We use similar hyper-parameters as above, but that we increase the hidden layer size to 100 
hidden units. As before we outperform the state of the art while training a single model 
that is able to generalize to new sequence lengths.

\subsubsection*{Random permutation problem}
In this case we have a dictionary of 100 symbols. Except the first and last position which have 
the same value  
sampled from $\{1, 2 \}$ the other entries are randomly picked from $[3, 100]$. 
The task is to do next symbol prediction, though the only predictable symbol is 
the last one. 

We use a 100 hidden units with a learning rate of .001 and $\alpha$, the regularization coefficient, set to 1. The cutoff 
threshold is left to 6. This task turns out to be more difficult to learn, and only 1 out 
of 8 experiments succeeded. As before we use a single model to deal with multiple 
values for $T$ (from 50 to 200 units). 

\subsubsection*{Noiseless memorization problem}

For the noiseless memorization we are presented with a binary pattern of length 5, followed 
by $T$ steps of constant value. After these $T$ steps the model needs to generate the 
pattern seen initially. We also consider the extension of this problem from
\citet{Martens+Sutskever-ICML2011}, where the pattern has length 10, and the symbol set 
has cardinality 5 instead of 2. 

We manage a 100\% success rate on these tasks, though we train a different model for the 
5 sequence lengths considered (50, 100, 150, 200). 

\subsection*{Natural Tasks}
\subsubsection*{Polyphonic music prediction}

We train our model, a sigmoid units RNN, on sequences of 200 steps.  The cut-off coefficient threshold 
is the same in all cases, namely 8 (note that one has to take the mean over the sequence length when 
computing the gradients). 

In case of the Piano-midi.de dataset we use 300 hidden units and an initial learning rate of 1.0 (whir the learning 
rate halved every time the error over an epoch increased instead of decreasing).
For the regularized model we used a initial value for regularization coefficient $\alpha$ of 0.5, where $\alpha$ follows a $1/t$ schedule, i.e. 
$\alpha_t = \frac{1}{2t}$ (where $t$ measures the number of epochs). 

For the Nottingham dataset we used the exact same setup. For MuseData we increased the hidden layer to 400 hidden units.
The learning rate was also decreased to 0.5. 
For the regularized model, the initial value for $\alpha$ was 0.1, and $\alpha_t = \frac{1}{t}$. 

We make the observation that for natural tasks it seems useful to use a schedule that decreases the regularization term.
We assume that the regularization term forces the model to focus on long term correlations at the expense of short term 
ones, so it may be useful to have this decaying factor in order to allow the model to make better use of the short term information.

\subsubsection*{Language modelling}

For the language modelling task we used a 500 sigmoidal hidden units model with no biases \citep{TomasIlya}. 
The model is trained over sequences of 200 steps, where the hidden state is carried over from one step to the next one. 

We use a cut-off threshold of 45 (though we take the sum of the cost over the sequence length) for all experiments. 
For next character prediction we have a learning rate of 0.01 when using clipping with no regularization term, 
0.05 when we add the regularization term and 0.001 when we do not use clipping. 
When predicting the 5th character in the future we use a learning rate of 0.05 with the regularization term and 0.1 without it. 

The regularization factor $\alpha$ for next character prediction was set to .01 and kept constant, while for the modified task 
we used an initial value of 0.05 with a $\frac{1}{t}$ schedule.

\end{document}